\documentclass{article}
\usepackage{pgfplots}
\usepackage{tabularray}
\usepackage{multirow}
\usepackage[final]{neurips_2023}
\usepackage{fancyvrb}
\usepackage{subcaption}

\usepackage[utf8]{inputenc} % allow utf-8 input
\usepackage[T1]{fontenc}    % use 8-bit T1 fonts
\usepackage{hyperref}       % hyperlinks
\usepackage{url}            % simple URL typesetting
\usepackage{booktabs}       % professional-quality tables
\usepackage{amsfonts}       % blackboard math symbols
\usepackage{nicefrac}       % compact symbols for 1/2, etc.
\usepackage{microtype}      % microtypography
\usepackage{xcolor}         % colors

\newcommand{\calico}[0]{\textsc{Calico}}
\newcommand{\modelName}[0]{\calico{}}
\newcommand{\linguist}[0]{\textsc{Linguist}}

\title{\modelName{}
    \makebox[0pt][l]{\includegraphics[width=0.05\textwidth]{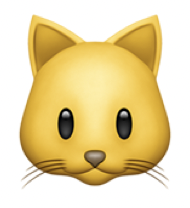}}
    \hspace{0.6cm}:
Conversational Agent Localization via Synthetic Data Generation}

\makeatletter
\newcommand{\myfnsymbol}[1]{%
  \expandafter\@myfnsymbol\csname c@#1\endcsname
}
\newcommand{\@myfnsymbol}[1]{%
  \ifcase #1
  \or 1% 1
  \or 2% 2
  \or \TextOrMath{\textasteriskcentered}{*}% 3
  \or \TextOrMath{\textdagger}{\dagger}% 4
  \fi
}
\newcommand{\affiliationA}{\@myfnsymbol{1}}
\newcommand{\affiliationB}{\@myfnsymbol{2}}
\newcommand{\equalcontributor}{\@myfnsymbol{3}}
\newcommand{\correspondingA}{\@myfnsymbol{4}}
\makeatother

\author{%
  Andy Rosenbaum\textsuperscript{\affiliationA}\\
  \texttt{andros@amazon.com} \\
  \And
  Pegah Kharazmi\textsuperscript{\affiliationA}\\
  \texttt{pkkharaz@amazon.com} \\
  \And
  Ershad Banijamali\textsuperscript{\affiliationA}\\
  \texttt{ebanijam@amazon.com} \\
  \And
  Lu Zeng\textsuperscript{\affiliationA}\\
  \texttt{luzeng@amazon.com} \\
  \And
  Chris DiPersio\textsuperscript{\affiliationA}\\
  \texttt{dipersio@amazon.com} \\
  \And
  Pan Wei\textsuperscript{\affiliationA}\\
  \texttt{panwei@amazon.com} \\
  \And
  Gokmen Oz\textsuperscript{\affiliationA}\\
  \texttt{ogokmen@amazon.de} \\
  \And
  Clement Chung\textsuperscript{\affiliationB}\\
  \texttt{chungcle@amazon.com} \\
  \And
  Karolina Owczarzak\textsuperscript{\affiliationB}\\
  \texttt{karowc@amazon.com} \\
  \And
  Fabian Triefenbach\textsuperscript{\affiliationB}\\
  \texttt{triefen@amazon.de} \\
  \And
  Wael Hamza\textsuperscript{\affiliationB}\\
  \texttt{waelhamz@amazon.com} \\
  \AND
  \textnormal{Amazon, Alexa AI}
}

\begin{document}

\maketitle

\footnotetext[1]{Individual Contributor}%
\footnotetext[2]{Manager/Advisor}%
\setcounter{footnote}{0}% Restart footnote counter
% Footnotes for rest of document uses \fnsymbol (or whatever you choose)
\renewcommand{\thefootnote}{\fnsymbol{footnote}}

\begin{abstract}
We present \modelName{}, a method to fine-tune Large Language Models (LLMs)
to localize conversational agent training data from one language to another.
For slots (named entities), \modelName{} supports three operations: verbatim copy,
literal translation, and \textit{localization},
i.e. generating slot values more
appropriate in the target language,
such as city and airport names located in countries where the language is spoken.
Furthermore, we design an iterative filtering mechanism to discard noisy generated samples, which we show boosts the performance of the downstream conversational agent. To prove the effectiveness of \modelName{}, we build and release a new
human-\textit{localized} (HL) version of
the MultiATIS++ travel information test set in 8 languages.
Compared to the original human-translated (HT) version of the test set,
we show that our new HL version is more challenging.
We also show that \modelName{} out-performs state-of-the-art \linguist{} (which relies on literal slot translation out of context)
both on the HT case, where \modelName{} generates more accurate slot translations,
and on the HL case, where \modelName{} generates \textit{localized} slots
which are closer to the HL test set.

\end{abstract}

\section{Introduction and Related Work}

Conversational agents usually apply Intent Classification (IC) and Slot Tagging (ST)
(also known as Named Entity Recognition or NER) to infer semantics from the
text of an agent-directed request \citep{TurSLU2011}.
In order to support global user bases, these agents are often multilingual.

IC+ST training data is typically scarce, especially in multilingual settings.
While Large Langauge Models (LLMs) can perform IC+ST from few examples (e.g. \citet{parikh-etal-2023-exploring}),
lightweight models such as encoder-only Transformers \citep{chen2019bert,xu20matis} are still useful
for cost- and latency-sensitive applications that support very high throughput.

Synthetic Data Generation (SDG) from Large Language Models (LLMs) has become a popular trend to address the data scarcity problem \citep{rosenbaum-llms-sdg-2023}.
SDG approaches relevant to the IC and ST tasks include back-translation, 
\citep{Bannard2005,sennrich-etal-2016-improving,edunov-etal-2018-understanding,Xie2020uda}
paraphrasing \citep{kumar-etal-2020-data,cho-etal-2019-paraphrase,malandrakis-etal-2019-controlled,Jolly2020,convai21-paraphrase-generation} 
word replacement \citep{zhang-etal-2020-seqmix,dai-adel-2020-analysis,wei-zou-2019-eda},
and carrier phrase regeneration \citep{kumar-etal-2022-controlled}.
A related thread is in-context generation of multilingual semantic parsing data \citep{rosenbaum-etal-2022-clasp} and
multi-party dialogs \citep{chen-etal-2023-places}.

\begin{figure}[t!]
\begin{Verbatim}[fontsize=\scriptsize, frame=single, commandchars=\\\{\}]
\textbf{INPUT:}
\textcolor{red}{<language>} \textcolor{teal}{\textbf{German}} \textcolor{red}{</language>}
\textcolor{red}{<domain>} \textcolor{black}{\textbf{travelinfo}} \textcolor{red}{</domain>}
\textcolor{red}{<intent>} \textcolor{violet}{\textbf{flight}} \textcolor{red}{</intent>}
\textcolor{red}{<include>}
 \colorbox{yellow!50}{[1 translation( first class ) ]} , 
 \colorbox{cyan!20}{[2 \underline{localization}( united airlines ) ]} , 
 \colorbox{orange!20}{[3 \underline{localization}( denver ) ]} , 
 \colorbox{green!20}{[4 \underline{localization}( baltimore ) ]} , 
 \colorbox{purple!20}{[5 translation( december seventeenth ) ]}
\textcolor{red}{</include>}
\textcolor{red}{<labels>}
 \colorbox{yellow!50}{[1=fare_class} , \colorbox{cyan!20}{[2=airline} , 
 \colorbox{orange!20}{[3=fromloc.city_name} , 
 \colorbox{green!20}{[4=toloc.city_name} , \colorbox{purple!20}{[5=date}
\textcolor{red}{</labels>}
\textcolor{red}{<examples>}
i need a \colorbox{yellow!50}{[1 first class ]} ticket
  on \colorbox{cyan!20}{[2 united airlines ]}
  from \colorbox{orange!20}{[3 denver ]}
  to \colorbox{green!20}{[4 baltimore ]}
  scheduled for \colorbox{purple!20}{[5 december seventeenth ]}
\textcolor{red}{</examples>}

\textbf{OUTPUTS}:
1. ich brauche ein \colorbox{yellow!50}{[1 erste klasse ]} ticket
   mit \colorbox{cyan!20}{[2 \underline{lufthansa} ]}
   von \colorbox{orange!20}{[3 \underline{hamburg} ]}
   nach \colorbox{green!20}{[4 \underline{köln} ]}
   für den \colorbox{purple!20}{[5 siebzehnten dezember ]}
\end{Verbatim}
\vspace{-.3cm}
\caption{
\modelName{} follows instructions to either literally translate entities 
(\texttt{fare\_class}, \texttt{date}) or to \textit{localize} entities
(\texttt{airline}, \texttt{from\_city}, \texttt{to\_city}).
}
\vspace{-.5cm}
\label{fig:prompt}
\end{figure}

In terms of \textit{cross-lingual} SDG for IC+ST,
Machine Translation with Slot Alignment (MT-SA) is a strong baseline \citep{xu20matis}, however the separate alignment step \textit{a posteriori}
introduces noise, which negatively impacts the quality of the generated data and the downstream task model.

Recently proposed \linguist{} \citep{rosenbaum-etal-2022-linguist} avoids the alignment problem by
first machine-translating the slot values
(out-of-context entities like ``new orleans'' or ``december sixteenth''),
and then generating a slot-annotated utterance in the target language incorporating
the machine-translated slot values.
% \footnote{
(An example of \linguist{} output is
``\texttt{book a flight to [1 new orleans ] on [2 december sixteenth ]}'', where \texttt{1} and \texttt{2} indicate slot labels \texttt{to\_city} and \texttt{date} respectively.
)
% }

% In this work, we make three core contributions: (1) propose the \modelName{} model for cross-lingual SGD of IC/NER training data, which resolves two important limitations of \linguist{} and shows significant improvements in downstream task accuracy, (2) 

In this work, we propose \modelName{} for cross-lingual SDG of IC+ST training data, which resolves two important limitations of \linguist{} (see Figure \ref{fig:prompt}):

(i) \textbf{Contextualized Slot Value Translation}:
\linguist{} translates the slots \textit{a priori} and out of context, which can lead to cascading errors,
due to the translation model choosing the wrong grammatical form in morphologically inflected languages,
or choosing the wrong semantic translation altogether. (For example, ``light'' can be a noun, synonym of ``lamp''; or an adjective, opposite of ``heavy''; or a verb.)
By contrast, \modelName{} translates the slot values and carrier phrase text jointly,
while producing the same
slot-annotated output format as \linguist{} to avoid the alignment problem of MT-SA.

(ii) \textbf{Slot Value Localization}: in real-world systems,
users are more likely to ask for entities specific to their locale,
e.g. in German booking flights to or from ``köln'' on ``lufthansa'' instead
of just asking for English entities and their translations, like ``denver'' or a literal 
translation of ``united airlines''. \modelName{} introduces
a \texttt{localization} operator which instructs the model to replace the value
in the source language with
a \textit{localized} version of the slot while translating the rest of the text around it.

\renewcommand{\thefootnote}{\arabic{footnote}}
To demonstrate the utility of slot value \textit{localization},
we create a new human-localized (HL) version of the MultiATIS++ test set in all 9 languages\footnote[1]{https://catalog.ldc.upenn.edu/LDC2021T04}, and benchmark
\modelName{} compared to \linguist{} on the 6 languages shared with our data generation model.
We show that our HL test set is more challenging that the original human-translated
(HT) test set.
We also show that \modelName{} out-performs \linguist{} both on the original HT
version, by producing more accurate slot translations with the full sentence context,
and on the new HL version by producing more relevant training data with localized slot values
like city and airport names.

Furthermore, we improve the process of selecting from among the n-best generated \modelName{} outputs:
instead of taking the output with lowest perplexity,
we design an Iterative Filtering Mechanism (IFM)
inspired by data augmentation through weak supervision 
\citep{Chen2022}.
We use the downstream task model
(IC+ST encoder fine-tuned on real data plus selected \modelName{}-generated synthetic data)
to \textit{re-select from among the n-best outputs based on matching the
intent and slots} requested in the prompt.
We show that the IFM
improves the final IC+ST performance on the
MultiATIS++ test set.

In summary, our contributions are
threefold:
(1) We propose \modelName{} to localize IC+ST training data with controls to either copy, translate, or \textit{localize} slot values;
(2)
we create a new version of the MultiATIS++ non-English test set, which includes updated text 
and annotation with human-\textit{localized} slot values such as city and airport names,
benchmark our models on it, and release the test set;
and
(3) we design an iterative filtering mechanism to select model generated data and show that it improves IC+ST performance on the
MultiATIS++ test set (both original and human-localized versions)
compared to selecting the output with lowest perplexity.

\vspace{-0.4cm}
\section{Methodology}
\vspace{-0.2cm}
Like \linguist{}, \modelName{} is a generative Large Language Model (LLM)
fine-tuned on an instruction prompt to generate synthetic training data
for IC+ST. \modelName{} takes inspiration from the \linguist{}
prompt, and supports additional slot operations.

\subsection{\modelName{} Prompt Design}

The \modelName{} prompt (Figure \ref{fig:prompt}) differs from \linguist{}
by adding controls at the slot level for three operations:
\texttt{unchanged}, indicating a verbatim copy (e.g. for flight numbers),
literal \texttt{translation}, and \texttt{localization},
i.e. replacement with a value more appropriate in the target language.

The \texttt{translation} operation of \modelName{} improves the slot translation quality compared to
out-of-context MT applied \textit{a priori} to \linguist{}.
For example, without any context, the word ``second'' in English could be reasonably translated to Spanish as either 
``\textit{segundo}'', ``\textit{segunda}'', ``\textit{segundos}'', or ``\textit{segundas}'',
depending on the plurality and grammatical gender of the Spanish noun it modifies.
Furthermore, if ``second'' is part of the phrase ``the second of september'', then it should be translated as ``\textit{dos}'', meaning ``two''.
\modelName{} attends to the entire input English sentence when generating translated values,
and therefore can disambiguate such cases.

\vspace{-.2cm}
\subsection{Training the \modelName{} Model}

Similar to \linguist{}, we fine-tune \modelName{} from AlexaTM 5B seq2seq
on cross-lingual prompts extracted from 
MASSIVE \citep{jgmf22massive}.
See details in Section \ref{sec:massive}. Models are finetuned for 10 epochs using a batch size of 16. 

\begin{figure}[h!]
  \centering
  \vspace{-.2cm}
  \includegraphics[width=.7\linewidth]{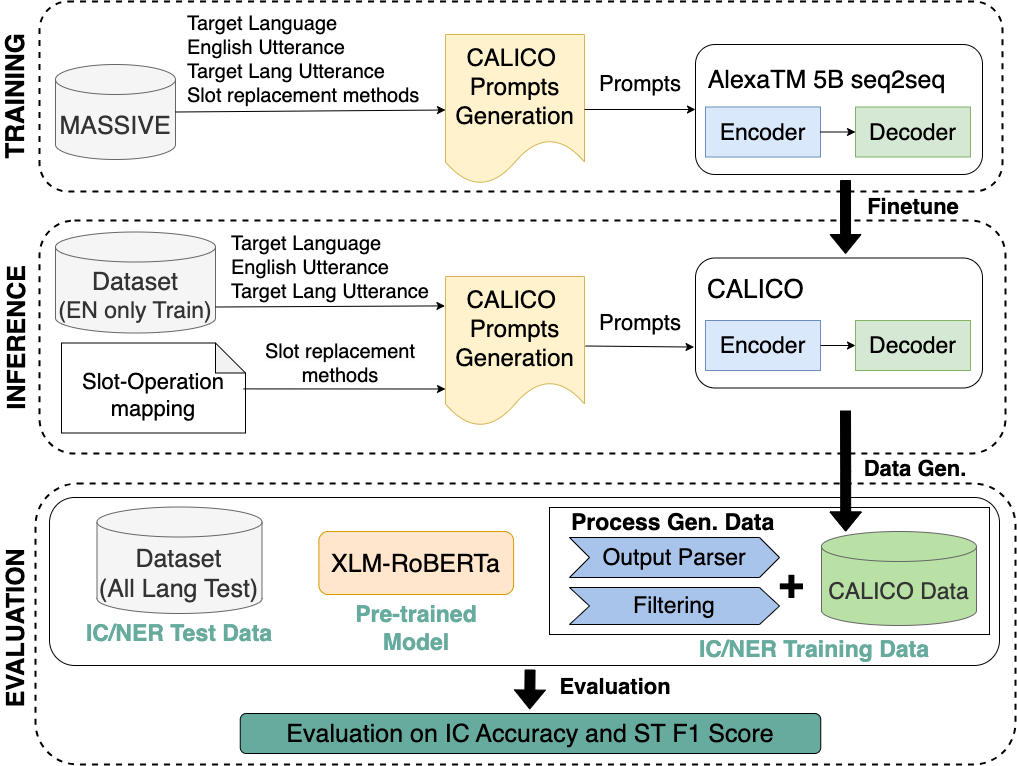}
  \caption{\modelName{} Training/Inference/Evaluation}
  \label{fig:methods}
  \vspace{-.5cm}
\end{figure}

\vspace{-.3cm}
\subsection{Inference }
For the target dataset, we take the English training data and instruct
\modelName{} to generate corresponding data in the target language.
We sample 8 outputs with top\_p 0.95 and temperature 1.0 and then filter down to select only one as described next.

\vspace{-0.2cm}
\subsection{Iterative Filtering Mechanism (IFM)}

We extend the post-processing pipeline of \linguist{}
to include an Iterative Filtering Mechanism (IFM) to select higher
quality samples from among the n-best \modelName{}-generated outputs.

Similar to \linguist{}, we first discard any outputs that do not pass heuristic string-based validation such as missing or extra brackets.
We then apply English-IC filtering and backoff to the original English example in case no output is valid, in order to maintain the original per-intent distribution.

In the first round, we randomly select one of the remaining \modelName{} outputs and fine-tune the IC+ST task model on real English data
plus the selected \modelName{} outputs in the other languages.

Then, we \textit{re-select from among the \modelName{} n-best outputs}: we discard samples where the IC+ST model hypothesis disagrees
with the intent and slot labels prompted for, then
randomly select a single output again, using a different random
seed to ensure that we don't always re-select the same output.

We apply this IFM repeatedly until performance plateaus (two iterations in all of our experiments).

\vspace{-.2cm}
\vspace{-.1cm}
\section{Experiments}

\vspace{-.1cm}
\subsection{Models}
\vspace{-.1cm}

We fine-tune AlexaTM 5B seq2seq model \citep{rosenbaum-etal-2022-linguist,Soltan2022AlexaTM2F,FitzGerald2022} as the \modelName{} 
data generaiton model.
For the downstream task IC+ST model and iterative filtering model, we fine-tune \texttt{xlm-roberta-base} \citep{conneau-etal-2020-unsupervised} (12 layers, 768 hidden dimension),
from the HuggingFace \citep{wolf-etal-2020-transformers} implementation.

\vspace{-.3cm}
\subsection{Datasets}

We fine-tune \modelName{} on cross-lingual prompts extracted from MASSIVE,
which contains parallel IC+ST annotated
data in 51 languages.
We fine-tune on 6 languages:
German, Spanish, French, Hindi, Japanese, and Portuguese,
each parallel to English.

After the \modelName{} data generation model is trained, we apply it on two IC+ST datasets,
MultiATIS++ and MultiSNIPS, to localize training data from English into the target languages.
The \modelName{} model has never seen the specific
intent and slot names, annotation scheme, or data conventions of the target downstream tasks,
and therefore must generalize at inference.

\subsubsection{MASSIVE}
\label{sec:massive}

MASSIVE \citep{jgmf22massive}, Multilingual Amazon SLURP (SLU resource package) for Slot Filling, Intent Classification, and Virtual-Assistant Evaluation, contains 19,521 English realistic, labeled virtual-assistant utterances spanning 18 domains, 60 intents, and 55 slots. It is a parallel dataset, where each English utterance is localized or translated into 50 typologically diverse languages. \textit{The dataset includes annotations on the human-chosen replacement method for each slot} (i.e. translation or localization or unchanged) for each pair of English and parallel target language utterance. Crucially, we use these slot-level annotations in the
prompts for \modelName{} fine-tuning so that the model learns
to follow instructions like \texttt{localization} (Figure \ref{fig:prompt}).

\begin{table*}[b!]
\vspace{-.3cm}
\footnotesize
\centering
\begin{tabular}{c| c c|c c c c c}
\hline
\multirow{2}{*}{\textbf{\shortstack{Lang}}}
& \multicolumn{7}{|c}{\textbf{\shortstack{Eval on test data with HT slots}}} \\
\cline{2-8}
& \textbf{\shortstack{Lower\\bound}}  
& \textbf{\shortstack{Upper\\bound}}  
& \textbf{\shortstack{\linguist{}\\(our repro)}}
& \textbf{\shortstack{\modelName{}\\(No IFM)}}
& \textbf{\shortstack{\modelName{}  \\  (IFM)}}
& \textbf{\shortstack{\modelName{} \\(All Transl.)}}
& \textbf{\shortstack{IFM \\ comb all}}
\\

\hline
& \multicolumn{7}{|c}{IC Accuracy (\%)}  \\
\hline
EN & 97.10   &   97.54 & 97.77 & 97.66 & 97.77 & \textbf{97.99} &   \underline{97.88}  \\
\hline
DE & 90.96   &   97.43 & 94.20 & \textbf{97.10} & \textbf{97.10} &   \textbf{97.10} &   \underline{96.88} \\
ES & 95.09   &   96.76 & 96.76 & 97.32 & \textbf{97.77} & 97.43 & \underline{97.54} \\
FR & 94.08   &   97.88 & 96.99 & 97.88 & 97.66 & \underline{97.88} &  \textbf{98.21}\\
HI & 85.38   &   94.64 & 90.62 & 92.41 & \textbf{95.31} & 92.30 &      \underline{94.75}\\
JA & 87.61   &   96.28 & 89.30 & 94.48 & \textbf{96.85} & {95.05} &     \underline{96.51}\\
PT & 91.63   &   96.88 & 96.21 & 95.98 & \textbf{97.10} & 96.32 &     \underline{96.65 }\\
\hline
AVG non-EN     & 90.79  & 96.65 & 94.01 & 95.86 & \textbf{96.96} & {96.01} &  \underline{96.76} \\
\hline
\hline
& \multicolumn{7}{|c}{ST F1} \\
\hline
EN &  95.73   &   96.01 & 95.69 & 95.71 & \textbf{95.85} & \underline{95.76} &    95.68 \\
\hline
DE &  80.47   &   95.34 & 84.11 & 87.47 & \underline{87.68} & 85.76 &    \textbf{89.23} \\
ES &  81.53   &   87.55 & 84.43 & 86.81 & \textbf{88.32} & {87.26}  &   \underline{87.57} \\
FR &  77.69   &   94.56 & 84.99 & \underline{85.71} & \textbf{86.59} & 85.10 &     85.70 \\
HI &  64.45   &   87.81 & 77.88 & 76.92 & 76.32 & \underline{82.69} &   \textbf{83.04} \\
JA &  41.98   &   93.64 & 85.15 & 88.43& 90.71  & \underline{91.85} & \textbf{92.38} \\
PT &  50.50   &   92.16 & \textbf{84.66} & 81.90 & \underline{83.83} & 81.06 &     83.12 \\
\hline
AVG non-EN     & 66.10  & 91.84 & 83.54 & 84.54 & 85.58 & \underline{85.62} &    \textbf{86.84}\\
\hline
\end{tabular}
\caption{\footnotesize{MultiATIS++ data generation performance on IC accuracy and ST F1 Score.
The result is reported on Human-Translate (``HT'') test set. The best (second best) result is bold (underlined).}}
\label{tab:matisppResHT}
\end{table*}
\vspace{-0.4cm}

\begin{table*}[b!]
\footnotesize
\centering
\begin{tabular}{c| c c|c c c c c}
\hline
\multirow{2}{*}{\textbf{\shortstack{Lang}}}
& \multicolumn{7}{|c}{\textbf{\shortstack{Eval on test data with HL slots}}} \\
\cline{2-8}
& \textbf{\shortstack{Lower\\bound}}  
& \textbf{\shortstack{Upper\\bound}}  
& \textbf{\shortstack{\linguist{}\\(our repro)}}
& \textbf{\shortstack{\modelName{}\\(No IFM)}}
& \textbf{\shortstack{\modelName{}  \\  (IFM)}}
& \textbf{\shortstack{\modelName{} \\(All Transl.)}}
& \textbf{\shortstack{IFM \\ comb all}} \\

\hline
& \multicolumn{7}{|c}{IC Accuracy (\%)}  \\
\hline
EN & 97.10  &     97.54     &   97.77 &	97.66 &	 97.77 & \textbf{97.99} & \underline{97.88}\\
\hline
DE &  90.40  &    97.10     &   94.42 &	        96.88 & \textbf{97.21} &	 \underline{97.10}  & {96.99}\\
ES & 95.54  &    96.43     &   96.88 &	\underline{96.88}& 96.76  &	 96.76  & \textbf{96.99} \\
FR & 93.19  &    96.76     &   96.21 &	\underline{96.76}& 96.21  &	 96.65   & \textbf{96.88}\\
HI &  86.16  &    93.08     &   91.29 &	        {92.86} & \underline{94.98} &	 93.53  & \textbf{95.31}\\
JA & 86.38  &    96.09     &   89.29 &	{94.53} & \textbf{95.98} &	 94.98   & \underline{95.31}\\
PT & 91.52  &    96.09     &   95.54 &	95.20 & \underline{95.76} &	 95.54 & \textbf{95.98}\\
\hline
AVG non-EN     &   90.53  &    95.93     &    93.94 &	95.52& \underline{96.15}  &	 95.76 & \textbf{96.24}\\
\hline
\hline
& \multicolumn{7}{|c}{ST F1} \\
\hline
EN &  95.73   &    96.01     &    95.69 &	95.71 & \textbf{95.85} &	 \underline{95.76}  & 95.68\\
\hline
DE &  76.61   &    85.78     &   79.96 &	{80.36} & \underline{82.52} &	 78.97  & \textbf{83.54}\\
ES &  71.86   &    76.08     &   74.11 &	{83.57}& \textbf{85.83}  &	 {78.14} & \underline{84.50}\\
FR & 77.62   &    85.79     &   \underline{82.39} &	{82.07} & \textbf{83.10} &	 81.63   & 81.97\\
HI & 72.84   &    84.93     &   76.52 &	78.62 & \underline{80.79} &	 79.99& \textbf{81.17}\\
JA &  41.61   &   91.19      &   86.34 &	{89.53} & \textbf{91.61} &	 88.98  &\underline{91.39} \\
PT & 77.65   &    90.18     &   83.63 &	82.69 & \textbf{85.00} &	 81.25  &\underline{84.07} \\
\hline
AVG non-EN     &69.70   & 85.66     &   80.49 &	82.81 & \textbf{84.81} &	 81.49  & \underline{84.44} \\
\hline
\end{tabular}
\caption{\footnotesize{MultiATIS++ data generation performance on IC accuracy and ST F1 Score.
The result is reported on new localized version (``HL'') of the test set. The best (second best) result  is bold (underlined).}}
\label{tab:matisppResHL}
\end{table*}

\begin{table}[t!]
\centering
\footnotesize
\label{table:matispp_icner_performance}
\begin{tabular}{c|c c|c c c} % \begin{tabular}{| C{.8cm}| C{.45cm}  C{.45cm}|C{.8cm}   C{1cm}  C{.8cm}|}0.
\hline
\textbf{\shortstack{Lang}} 
& \textbf{\shortstack{Lower\\ bound}} 
& \textbf{\shortstack{Upper\\ bound}}  
& \textbf{\shortstack{\linguist{}}}
& \textbf{\shortstack{\modelName{}}}
& \textbf{\shortstack{\modelName{} \\(All Transl.)}}\\
\hline
& \multicolumn{5}{|c}{IC Accuracy (\%)} \\
\hline
EN &99.01 &98.86 & 98.86 & 99.01 & 98.79\\
\hline
ES &95.03 & 98.15 & 98.44 & 98.30 & 98.01\\
FR &96.59 & 98.58 & 98.72 & 98.44 & 97.94\\
HI &92.39 & 95.11 & 98.71 & 97.56  & 97.70\\
\hline
AVG non EN & 94.67 & 97.28 & 98.62 & 98.10 & 97.88 \\
\hline
\hline
& \multicolumn{5}{|c}{ST F1} \\
\hline
EN &93.63 & 95.48 & 95.24 & 95.35 & 95.11\\
\hline
ES &70.14 & 91.35 & 90.35 & 90.42 & 89.71\\
FR &70.47 & 86.73 & 84.35 & 88.81 & 88.53\\
HI &47.85 & 85.72 & 82.87 & 77.90 & 83.99\\
\hline
AVG non EN &62.82 & 87.93 & 85.86 & 85.71 & 87.41\\
\hline
\end{tabular}
\caption{\footnotesize{MultiSNIPS data generation performance on IC accuracy and ST F1 Score.}}
\label{tab:msnipsRes}
\vspace{-.8cm}
\end{table}
\vspace{.3cm}
\subsubsection{MultiATIS++}

MultiATIS++ dataset extends the English-only Air Travel Information Services dataset ATIS to 9 languages via human translation. Our work focuses on the 7 languages MultiATIS++ shares with our pre-trained model: English (EN), German (DE), Spanish (ES), French (FR), Hindi (HI), Japanese (JA), Portuguese (PT), with 4488 (HI: 1440) training utterances, 490 (HI: 160) validation utterances, and 893 test utterances, over 18 (HI: 17) intents and 84 (HI: 75) slots.
(The test set release also contains Turkish and Chinese.)

Since the original test set is human-translated from English, it contains only translated slot values (e.g. for ``new orleans'' in English, the Spanish version would be
``nueva orleans''), but not \textit{localized} entities.
To showcase the effectiveness of \modelName{}, we create a \textbf{new version of the MultiATIS++ test set} in all 8 non-English languages, by 
asking human experts to \textit{localize} the slot values for 8 slot types. Specifically, the human experts replace the original human-translated
English slot value (e.g. ``nueva orleans'' in Spanish)
with a value more appropriate value in the target language (e.g. ``madrid'').
The localized slot types are
\texttt{airline\_code}, \texttt{airline\_name}, \texttt{airport\_code}, \texttt{airport\_name}, \texttt{city\_name}, \texttt{country\_name}, \texttt{state\_code}, \texttt{state\_name}.

We also ask the experts to modify the carrier phrase text as needed to make the entire
request grammatically correct. For example, in French the experts might change the form
of the definite article ``le'', ``la'', ``les'', or ``l''' (all meaning ``the'') to
match the plurality, gender, and pronunciation of the newly chosen slot value.
The rest of the slot types and text remain as they are in the original.

\vspace{-0.3cm}
\subsubsection{MultiSNIPS}

The MultiSNIPS dataset \citep{eacl-2023-asa-sailik} contains
human translations of the SNIPS \citep{coucke2019efficient}
dataset into three languages: Spanish (ES), French (FR), and Hindi (HI).
\vspace{-0.3cm}

\subsection{Results}

\textbf{MultiATIS++ Results} are shown in Tables \ref{tab:matisppResHT} and \ref{tab:matisppResHL}, where the upper block reports IC Accuracy, and the lower block reports ST F1 Score.
Table \ref{tab:matisppResHT} is reported on the original test set,
which contains human-translated slot values, whereas Table \ref{tab:matisppResHL} is on our new version of the test set,
where the slots are human-localized to versions more common in the target language.

``Lower bound'' indicates the IC+ST model trained \textit{only on the English training data}.
``Upper bound'' indicates the IC+ST model trained on MultiATIS++ training data for all 7 languages. The remaining columns indicate IC+ST models trained on the concatenation of original English training data with synthetic training data for the other 6 languages from one or more methods.
``\linguist{} (our repro)'' is our reproduction of \linguist{}.

``\modelName{} (IFM)'' is our candidate approach, where we apply the \texttt{localization}, \texttt{translation}, and \texttt{copy} operations to specific slot values as shown in Table \ref{table:matispp_slot_ops} (Appendix \ref{sec:slot_ops}), along with post-generation iterative filtering mechanism. 
\iffalse
, as show in Table \ref{table:matispp_slot_ops} (Appendix \ref{sec:slot_ops}).
\fi
``\modelName{} (All Transl.)'' is our candidate model with the \texttt{translation}
operation applied to all slots at inference time.
In ``\modelName{} (No IFM)'', we do an ablation setup on the iterative filtering mechanism, where we do not perform iterative filtering. And finally in the column ``IFM Comb all'', we include the synthetic data from ``\modelName{} (IFM)'' combined with ``\linguist{} (our repro)'' and ``\modelName{} (All Translate)''.

We primarily focus on ``AVG non EN'', which is the average across the non-English languages. On the original test set (Table \ref{tab:matisppResHT}),
we find that ``\modelName{} (IFM)'' is the best performing single method on average, surpassing \linguist{} by 2.95 points absolute on IC (from 94.01 to 96.96) and 2.04 points absolute on ST F1 (from 83.54 to 85.58).

The improvement of \modelName{} over \linguist{} is reflected on most languages, however Japanese (JA) shows the most improvement, having +7.55 / +5.56 (from 89.30 to 96.85  / from 85.15 to 90.71) points improvement on IC / ST.
This suggests that Japanese was
being limited the most by the drawbacks of \linguist{}, e.g. making translation mistakes out of context.

On the new localized test data (Table \ref{tab:matisppResHL}), both versions of \modelName{} improve over \linguist{}, and IFM improves even further.
With \modelName{} plus IFM, we improve on IC from 93.94 \linguist{} to 96.15 (for 2.21 points absolute) and ST from 80.49 \linguist{} to 84.81 (i.e. 4.32 points absolute).
As with the HT test set, the improvement is particularly large for Japanese (JA).

On both settings, we combine data from LINGUIST and both versions of \modelName{}, however find that the gains are not consistently
synergistic.

We see the performance improvement of \modelName{} over \linguist{} is directly correlated with the ``Success Rate'' (Table \ref{tab:matisppSR} in Appendix \ref{sec:successRates}, filtering to keep only those outputs which pass string-matching heuristics and IC hypothesis filtering),
indicating that the data produced from \modelName{}
is cleaner and more usable than that of \linguist{}.

All data generation models and even the Upper Bound of including human-translated training data perform significantly worse on the test data with human-localized slots compared to the original human-translated test data,
indicating that the human-localized test set is more challenging,
and motivating future work on conversational agent localization.

\textbf{MultiSNIPS Results} are show in Table \ref{tab:msnipsRes}. Here, in the AVG non EN there are small 
differences overall compared to \linguist{}: \modelName{} (All Translate) is 0.74 points absolute worse on 
IC (from 98.62 to 97.88) and +1.55 points absolute better on ST (from 85.86 to 87.41).
However, similarly to the MultiATIS++ results, \modelName{} (All Translate) out-performs \linguist{}. All models are close to the upper bound, however, indicating that this dataset may not be particularly challenging.

\section{Conclusion and Future Work}
\vspace{-0.2cm}
We introduced \modelName{}, a novel pipeline for synthetic annotated data generation in new languages, via fine-tuning a largescale pre-trained multilingual seq2seq model. We demonstrated that unlike prior techniques that would translate slots out of context, \modelName{} can generate annotated slots based on the context and localize them with values more appropriate to the target language . In future, we plan to extend and leverage a reward model into a reinforcement learning setup to further improve the quality of the generated data. We would also like to explore ways to combine the positive effects of
\linguist{} paraphrasing with \modelName{} localization.

\bibliography{anthology,custom}
\bibliographystyle{acl_natbib}

\newpage
\section*{Appendix}
\appendix

\section{Prompt Comparison}
\label{sec:prompt_comparison}

\begin{figure}[h!]
\centering
\begin{subfigure}{.42\textwidth}
  \centering
  % \hfill
\begin{Verbatim}[fontsize=\scriptsize, frame=single, commandchars=\\\{\}]
\textbf{INPUT:}
\textcolor{red}{<language>} \textcolor{teal}{\textbf{Spanish}} \textcolor{red}{</language>}
\textcolor{red}{<domain>} \textcolor{black}{\textbf{travelinfo}} \textcolor{red}{</domain>}
\textcolor{red}{<intent>} \textcolor{violet}{\textbf{flight}} \textcolor{red}{</intent>}
\textcolor{red}{<include>}
 \colorbox{yellow!50}{[1 \underline{translation}( pittsburgh ) ]} , 
 \colorbox{cyan!20}{[2 \underline{translation}( new york ) ]} , 
 \colorbox{purple!20}{[ 3 translation( 5 pm ) ]}
\textcolor{red}{</include>}
\textcolor{red}{<labels>}
 \colorbox{yellow!50}{[1=fromloc.city_name} ,
 \colorbox{cyan!20}{[2=toloc.city_name} , 
 \colorbox{purple!20}{[3=depart_time.time}
\textcolor{red}{</labels>}
\textcolor{red}{<examples>}
i need a flight
  from \colorbox{yellow!50}{[1 pittsburgh ]}
  to \colorbox{cyan!20}{[2 new york ]}
  leaving at \colorbox{purple!20}{[3 5 pm ]}
\textcolor{red}{</examples>}

\textbf{OUTPUTS}:

1. necesito un vuelo
   desde \colorbox{yellow!50}{[1 \underline{pittsburgh} ]}
   a \colorbox{cyan!20}{[2 \underline{nueva york} ]}
   saliendo a las \colorbox{purple!20}{[3 cinco de la tarde ]}
\end{Verbatim}
% \vspace{-0.2cm}
\caption{
The prompt with \texttt{translation} operation.
}
  \label{fig:sub1}
\end{subfigure}%
\hspace{0.05 \textwidth}
\begin{subfigure}{.42\textwidth}
  \centering
\begin{Verbatim}[fontsize=\scriptsize, frame=single, commandchars=\\\{\}]
\textbf{INPUT:}
\textcolor{red}{<language>} \textcolor{teal}{\textbf{Spanish}} \textcolor{red}{</language>}
\textcolor{red}{<domain>} \textcolor{black}{\textbf{travelinfo}} \textcolor{red}{</domain>}
\textcolor{red}{<intent>} \textcolor{violet}{\textbf{flight}} \textcolor{red}{</intent>}
\textcolor{red}{<include>}
 \colorbox{yellow!50}{[1 \underline{localization}( pittsburgh ) ]} , 
 \colorbox{cyan!20}{[2 \underline{localization}( new york ) ]} , 
 \colorbox{purple!20}{[ 3 translation( 5 pm ) ]}
\textcolor{red}{</include>}
\textcolor{red}{<labels>}
 \colorbox{yellow!50}{[1=fromloc.city_name} ,
 \colorbox{cyan!20}{[2=toloc.city_name} , 
 \colorbox{purple!20}{[3=depart_time.time}
\textcolor{red}{</labels>}
\textcolor{red}{<examples>}
i need a flight
  from \colorbox{yellow!50}{[1 pittsburgh ]}
  to \colorbox{cyan!20}{[2 new york ]}
  leaving at \colorbox{purple!20}{[3 5 pm ]}
\textcolor{red}{</examples>}

\textbf{OUTPUTS}:

1. necesito un vuelo
   desde \colorbox{yellow!50}{[1 \underline{madrid} ]}
   a \colorbox{cyan!20}{[2 \underline{barcelona} ]}
   saliendo a las \colorbox{purple!20}{[3 cinco de la tarde ]}
\end{Verbatim}
% \vspace{-0.2cm}
\caption{
The prompt with \texttt{localization} operation.
}
  \label{fig:sub2}
\end{subfigure}
% \vspace{-0.2cm}
\caption{
For the same input, \modelName{} can follow the instruction prompt to
map city names from English into Spanish either via
literal translations (left)
(``pittsburgh'' $\rightarrow{}$ ``pittsburgh''
and ``new york'' $\rightarrow{}$ ``nueva york'')
or via \textit{localization} (right)
(``pittsburgh'' $\rightarrow{}$ ``madrid''
and ``new york'' $\rightarrow{}$ ``barcelona'').
}
\label{fig:output_zelda}
\end{figure}

\newpage
\section{Success Rates}
\label{sec:successRates}

\begin{table}[h!]
\caption{MultiATIS++ data generation Success rate.}
\centering
\label{table:matispp_success_rate}
\resizebox{\columnwidth}{!}{\begin{tabular}{|c|c|ccc|cc|c|}
\hline
\multicolumn{1}{|c}{\textbf{\shortstack{Method}}} & \multicolumn{1}{|c}{\textbf{\shortstack{Lang}}} &
\multicolumn{1}{|c}{\textbf{\shortstack{Parse\\Success (\%)}}} &
\multicolumn{1}{c}{\textbf{\shortstack{IC Filter\\Success (\%)}}} &
\multicolumn{1}{c}{\textbf{\shortstack{Final\\Success (\%)}}} &
\multicolumn{1}{|c}{\textbf{\shortstack{\# Utt. from\\Generation}}} &
\multicolumn{1}{c|}{\textbf{\shortstack{\# EN Utt.\\Copied}}} &
\multicolumn{1}{c|}{\textbf{\shortstack{Total}}} \\
\hline
\multirow{7}{*}{\textbf{\shortstack{\linguist{}\\(our repro)}}}&DE &91.61	&78.76	&72.15	&3037	&1172	&4209\\
&ES	&91.59	&80.54	&73.77	&3105	&1104	&4209\\
&FR	&92.47	&73.79	&68.23	&2872	&1337	&4209\\
&HI	&89.69	&79.00	&70.85	&2982	&1227	&4209\\
&JA	&72.18	&76.88	&55.49	&2336	&1873	&4209\\
&PT	&92.33	&83.30	&76.91	&3237	&972	&4209\\
\cline{2-8}
&AVG& 88.31	&78.71	&69.57	&2928	&1281	&4209 \\
\hline
\hline

\multirow{7}{*}{\textbf{\shortstack{\modelName{}}}}
    & DE& 99.44& 	90.78& 	90.27& 	4052& 	436& 	4488 \\
& ES& 	99.44& 	94.25& 	93.72& 	4206& 	282& 	4488\\
& FR& 	99.84& 	94.14& 	93.99& 	4218& 	270& 	4488\\
& HI& 	97.82& 	85.16& 	83.30& 	3739& 	749& 	4488\\
& JA& 	95.10& 	79.48& 	75.58& 	3392& 	1096& 	4488\\
& PT& 	99.31& 	92.74& 	92.10& 	4133& 	355& 	4488\\
\cline{2-8}
&AVG& 98.49	& 89.43& 	88.16& 	3957& 	531& 	4488\\
\hline
\hline

\multirow{7}{*}{\textbf{\shortstack{\modelName{} \\(All Translate)}}}&DE&	99.35&	91.35&	90.76&	4073&	415&	4488 \\
&ES	&98.93	&94.01	&93.00	&4174	&314	&4488\\
&FR	&99.64	&94.43	&94.09	&4223	&265	&4488\\
&HI	&97.08	&85.76	&83.26	&3737	&751	&4488\\
&JA	&94.45	&82.66	&78.07	&3504	&984	&4488\\
&PT	&98.95	&92.96	&91.99	&4128	&360	&4488\\
\cline{2-8}
&AVG& 98.07	&90.20&	88.53&	3973	&515	&4488\\
\hline

\end{tabular}}
\label{tab:matisppSR}
\end{table}

\begin{table}[h!]
\caption{MultiSNIPS data generation Success rate.}
\centering
\label{table:msnips_success_rate}
\resizebox{\columnwidth}{!}{\begin{tabular}{|c|c|ccc|cc|c|}
\hline
\multicolumn{1}{|c}{\textbf{\shortstack{Method}}} & \multicolumn{1}{|c}{\textbf{\shortstack{Lang}}} &
\multicolumn{1}{|c}{\textbf{\shortstack{Parse\\Success (\%)}}} &
\multicolumn{1}{c}{\textbf{\shortstack{IC Filter\\Success (\%)}}} &
\multicolumn{1}{c}{\textbf{\shortstack{Final\\Success (\%)}}} &
\multicolumn{1}{|c}{\textbf{\shortstack{\# Utt. from\\Generation}}} &
\multicolumn{1}{c|}{\textbf{\shortstack{\# EN Utt.\\Copied}}} &
\multicolumn{1}{c|}{\textbf{\shortstack{Total}}} \\
\hline
\multirow{4}{*}{\textbf{\shortstack{\linguist{}\\(our repro)}}}&ES	& 97.10	& 88.67	& 86.10	& 10846	& 1751	& 12597 \\
&FR	& 97.30	& 97.74	& 95.10	& 11980 &617 &12597 \\
&HI	&72.86 	&92.34 	&67.28 	&8475 	&4122 	&12507 \\
\cline{2-8}
&AVG&89.09 	&92.92 	&82.83 	&10434 	&2163 	&12597 \\
\hline\hline

\multirow{4}{*}{\textbf{\shortstack{\modelName{}}}}& ES &99.98 	&90.75 	&90.73 	&11278 	&1152 	&12430 \\
& FR&99.98 	&97.61 	&97.59 	&12130 	&300 	&12430 \\
& HI&99.73 	&92.66 	&92.41 	&11487 	&943 	&12430 \\
\cline{2-8}
&AVG&99.90 	&93.67 	&93.58 	&11632 	&798 	&12430 \\
\hline\hline

\multirow{4}{*}{\textbf{\shortstack{\modelName{} \\(All Translate)}}}& ES&99.98 	&90.96 	&90.94 	&11304 	&1126 	&12430 \\
& FR&99.99 	&97.45 	&97.44 	&12112 	&318 	&12430 \\
& HI&99.85 	&92.30	&92.16 	&11456 	&974 	&12430 \\
\cline{2-8}
&AVG&99.94 	&93.57 	&93.51 	&11624 	&806 	&12430 \\
\hline

\end{tabular}}
\end{table}

\newpage

\section{CALICO Slot Operations}
\label{sec:slot_ops}
\begin{table}[h]
\caption{mATIS++ CALICO Slot Operations}
\centering
\label{table:matispp_slot_ops}
\begin{tabular}{|cc|cc|}
\hline
\textbf{Slot} & \textbf{Operation} & \textbf{Slot} & \textbf{Operation} \\
\hline
 airline\_code                & copy         & flight\_mod                  & translation  \\
\textbf{ airline\_name}                & \textbf{localization} & flight\_number               & translation  \\
 airport\_code                & copy         & flight\_stop                 & translation  \\
 \textbf{airport\_name}                & \textbf{localization} & flight\_time                 & translation  \\
 arrive\_date.date\_relative  & translation  & fromloc.airport\_code        & copy         \\
 arrive\_date.day\_name       & translation  & \textbf{fromloc.airport\_name}        & \textbf{localization} \\
 arrive\_date.day\_number     & translation  & \textbf{fromloc.city\_name}           & \textbf{localization} \\
 arrive\_date.month\_name     & translation  & fromloc.state\_code          & copy         \\
 arrive\_date.today\_relative & translation  & \textbf{fromloc.state\_name}          & \textbf{localization} \\
 arrive\_time.end\_time       & translation  & meal                         & translation  \\
 arrive\_time.period\_mod     & translation  & meal\_code                   & copy         \\
 arrive\_time.period\_of\_day & translation  & meal\_description            & translation  \\
 arrive\_time.start\_time     & translation  & mod                          & translation  \\
 arrive\_time.time            & translation  & month\_name                  & translation  \\
 arrive\_time.time\_relative  & translation  & or                           & translation  \\
 booking\_class               & translation  & period\_of\_day              & translation  \\
 \textbf{city\_name}                   & \textbf{localization} & restriction\_code            & copy         \\
 class\_type                  & translation  & return\_date.date\_relative  & translation  \\
 compartment                  & translation  & return\_date.day\_name       & translation  \\
 connect                      & translation  & return\_date.day\_number     & translation  \\
 cost\_relative               & translation  & return\_date.month\_name     & translation  \\
 day\_name                    & translation  & return\_date.today\_relative & translation  \\
 day\_number                  & translation  & return\_time.period\_mod     & translation  \\
 days\_code                   & copy         & return\_time.period\_of\_day & translation  \\
 depart\_date.date\_relative  & translation  & round\_trip                  & translation  \\
 depart\_date.day\_name       & translation  & state\_code                  & copy         \\
 depart\_date.day\_number     & translation  & \textbf{state\_name}                  & \textbf{localization} \\
 depart\_date.month\_name     & translation  & stoploc.airport\_code        & copy         \\
 depart\_date.today\_relative & translation  & \textbf{stoploc.airport\_name}        & \textbf{localization} \\
 depart\_date.year            & translation  & \textbf{stoploc.city\_name}           & \textbf{localization} \\
 depart\_time.end\_time       & translation  & stoploc.state\_code          & copy         \\ 
 depart\_time.period\_mod     & translation  & time                         & translation  \\
 depart\_time.period\_of\_day & translation  & time\_relative               & translation  \\
 depart\_time.start\_time     & translation  & today\_relative              & translation  \\
 depart\_time.time            & translation  & toloc.airport\_code          & copy         \\
 depart\_time.time\_relative  & translation  & \textbf{toloc.airport\_name}          & \textbf{localization} \\
 economy                      & translation  & \textbf{toloc.city\_name}             & \textbf{localization} \\
 fare\_amount                 & translation  & \textbf{toloc.country\_name}          & \textbf{localization} \\
 fare\_basis\_code            & translation  & toloc.state\_code            & copy         \\
 flight                       & translation  & \textbf{toloc.state\_name}            & \textbf{localization} \\
 flight\_days                 & translation  & transport\_type              & translation  \\
\hline
\end{tabular}
\end{table}

\end{document}